\documentclass[sigconf]{acmart}

\AtBeginDocument{%
  \providecommand\BibTeX{{%
    \normalfont B\kern-0.5em{\scshape i\kern-0.25em b}\kern-0.8em\TeX}}}

\usepackage[noend]{algpseudocode}
\usepackage{subcaption}
\usepackage{algorithmicx,algorithm}
\usepackage{color,soul}
\usepackage{enumitem}
\usepackage{multirow}
\usepackage{booktabs}
\usepackage{makecell}
\usepackage{tabularx}

\begin{document}


\title{DNNAbacus: Toward Accurate Computational Cost Prediction for Deep Neural Networks}


%
\author{Lu Bai}
\affiliation{%
  \institution{Beijing Institute of Technology}
  \streetaddress{Zhongguancun Street, Haidian District}
  \city{Beijing}
  \country{China}
  \postcode{100081}
}
\email{bl@bit.edu.cn}

\author{Weixing Ji}
\authornotemark[1]
\affiliation{%
  \institution{Beijing Institute of Technology}
  \streetaddress{Zhongguancun Street, Haidian District}
  \city{Beijing}
  \country{China}
  \postcode{100081}
}
\email{jwx@bit.edu.cn}


\author{Qinyuan Li}
\affiliation{%
  \institution{Beijing Institute of Technology}
  \streetaddress{Zhongguancun Street, Haidian District}
  \city{Beijing}
  \country{China}
  \postcode{100081}
}
\email{1120180710@bit.edu.cn}


\author{Xilai Yao}
\affiliation{%
  \institution{Beijing Institute of Technology}
  \streetaddress{Zhongguancun Street, Haidian District}
  \city{Beijing}
  \country{China}
  \postcode{100081}
}
\email{3120211069@bit.edu.cn}


\author{Wei Xin}
\affiliation{%
  \institution{Alibaba-inc}
  \country{China}
}
\email{xinwei.xw@alibaba-inc.com}

\author{Wanyi Zhu}
\affiliation{%
  \institution{Alibaba-inc}
  \country{China}
}
\email{wanyi.wyz@alibaba-inc.com}


%

\renewcommand{\shortauthors}{Trovato and Tobin, et al.}
\renewcommand{\shortauthors}{}

\begin{abstract}
Deep learning is attracting interest across a variety of domains, including natural language processing, speech recognition, and computer vision. However, model training is time-consuming and requires huge computational resources. Existing works on the performance prediction of deep neural networks, which mostly focus on the training time prediction of a few models, rely on analytical models and result in high relative errors. 

This paper investigates the computational resource demands of 29 classical deep neural networks and builds accurate models for predicting computational costs. We first analyze the profiling results of typical networks and demonstrate that the computational resource demands of models with different inputs and hyperparameters are not obvious and intuitive. We then propose a lightweight prediction approach DNNAbacus with a novel network structural matrix for network representation. DNNAbacus can accurately predict both memory and time cost for PyTorch and TensorFlow models, which is also generalized to different hardware architectures and can have zero-shot capability for unseen networks. Our experimental results show that the mean relative error (MRE) is 0.9\% with respect to time and 2.8\% with respect to memory for 29 classic models, which is much lower than the state-of-the-art works.

\end{abstract}






\maketitle

\section{Introduction}
Deep neural network(DNN) has achieved considerable successes in recent years and is widely applied in many application areas due to massively available data and its powerful learning capability~\cite{2021survey}. There is no doubt that deep learning tasks have become the dominant workloads in data centers~\cite{naumov2020deep}. Intuitively, the performance of DNN can be improved using large datasets with more stacked layers and parameters. Consequently, the trend in deep learning is moving towards larger and deeper network designs~\cite{VDNN16}, which also increased the computation cost of the network. For example, a recurrent neural network with more than 1,000 layers is studied in \cite{he2016deep}. The large BERT model has 24 layers and 340 million parameters~\cite{DevlinCLT19}. Pre-training of BERT large takes four days on 4 to 16 Cloud TPUs, and each TPU has 64 GB of RAM~\cite{bertgithub}. 




 


Moreover, to accelerate the training and inference of deep neural models, graphics processing units (GPUs) and tensor processing units (TPUs) are designed and put into large-scale commercial deployment. Since the memory size required by the model is generally unknown beforehand, training tasks may fail due to insufficient memory on these devices. Studies have shown that a large proportion of deep learning job failures are caused by insufficient memory~\cite{ICSE2020MSRA}. Another study of 2,716 Stack Overflow posts also identified insufficient memory as one of the six main reasons for deep learning errors~\cite{FSE2019Bug}. The failure of model training tasks results in much more waste of computing and storage resources. 




In view of the above reasons, predicting the computational resources required by model training in advance is of great significance for resource allocation and task scheduling. Accurate predicting of execution time and maximum memory demands of model training can not only improve resource utilization but also reduce job failures in data centers. Recently, ever-increasing reports are working on the performance modeling and prediction~\cite{DNNMem2020}~\cite{perfnet2020}~\cite{PerfNet22021}~\cite{PerfEstimator2021}\cite{yang2021perfestimator}. However, most existing work focuses on training time prediction, and ignores the estimation of memory requirements, which may result in out-of-memory failure under trial-and-error method and arrive at the overheads finally charged to total throughput. Prior work either explores plainly features or builds an analytical model, resulting in a relatively high mean relative error. Moreover, a few models are used for evaluation in reported work, which is hard to present the generability of existing approaches. Though graph neural networks are also explored in recent work, suffered from high communication overhead and high relative error. 


In this paper, we profiled the training task of 29 typical networks on two platforms. We collected 17,300 data points from typical networks with different configurations and 5,500 data points from randomly generated deep neural networks. The execution time and maximum memory demands during network training are collected and investigated in detail. The results present that both time and memory required in model training are sensitive to inputs and hyperparameters. Even for the same model, training with different hyperparameters will lead to big differences in the demand for computing resources. This is because existing frameworks have different internal designs and invoke different convolutional algorithms at runtime. For example, PyTorch pre-allocates a large chunk of GPU memory and splits it into small blocks for fast reuse. It also uses a cache subsystem to improve the efficiency of GPU memory allocation and release. Altogether, it is a challenging task to predicate the computational resource demands for each configuration. Then we build a lightweight and accurate model DNNAbacus to predict the computational resource demands of training tasks based on an AutoML framework. Different from the previous analytical models, DNNAbacus estimates the training performance as a black-box model. It can accurately predict both memory and time cost for PyTorch and TensorFlow models. We also present a novel structural matrix based network representation technique, so that DNNAbacus is also generalized to different hardware architectures and can have zero-shot capability for unseen networks. We validate DNNAbacus using different datasets, computing platforms and frameworks. Our experimental results show that the mean relative error is about 0.9\% with respect to time and 2.8\% with respect to memory. Based on DNNAbacus and the genetic algorithm, we show the effectiveness of our approach in assigning training tasks to different machines.








\section{DNN Profiling and Analysis}

\subsection{Investigated Models and System Setup}
We profiled 29 deep neural networks in total, including VGG-16\cite{simonyan2014very}, GoogLeNet\cite{szegedy2015going}, ResNet-101\cite{he2016deep}, and ShuffleNet\cite{zhang2018shufflenet}. These are representative networks that have different structures. VGGNet stacks many $3\times3$ convolutional kernels. GoogLeNet assembles multiple convolutional and pooling operations into Inception modules. ResNet uses the residual module which shortens the connection between the front and back layers. MobileNet and SqueezeNet introduce depthwise separable convolution to structure lightweight deep neural networks, using a large number of $1\times1$ convolutional kernels to reduce the number of channels.
We collected and analyzed the total run time and maximum memory occupation of typical deep neural networks training with different hyperparameters on two datasets, MNIST and CIFAR-100. The investigated hyperparameters include data size, batch size, epoch, learning rate, optimizer, kernel size, padding, and stride. The ``time" package in Python is used to get the accurate run time. The ``pynvml" package is used to sample the GPU memory usage in real-time at 100 ms intervals and keep refreshing the maximum value for each model.
We tested the models on two systems and their setups are given in Table \ref{server}. 

\begin{table}[!ht]
    \caption{System Setup}
    \label{server}
    \begin{tabular}{c|c c}
    \hline
        Specification &  System 1 & System 2 \\ 
        \hline
        GPU Device & RTX2080 & RTX3090 \\ 
        \makecell{GPU Model} & Turing & Ampere \\ 
        GPU Memory & 11 GB & 24 GB \\ 
        CPU cores & 18 & 12 \\ 
        CPU Model & \makecell{i9-10980XE \\ 3.00 GHz} & \makecell{i9-10920X \\ 3.50 GHz} \\ 
        CUDA Version & 10.2 & 11.1 \\ 
        PyTorch Version & 1.8.1 & 1.8.1 \\
        TensorFlow Version & 1.15.0 & 1.15.0 \\
        \hline
    \end{tabular}
\end{table}

\subsection{Profiling Results}
\begin{figure}[ht]
\begin{center}
\centerline{\includegraphics[width=\columnwidth]{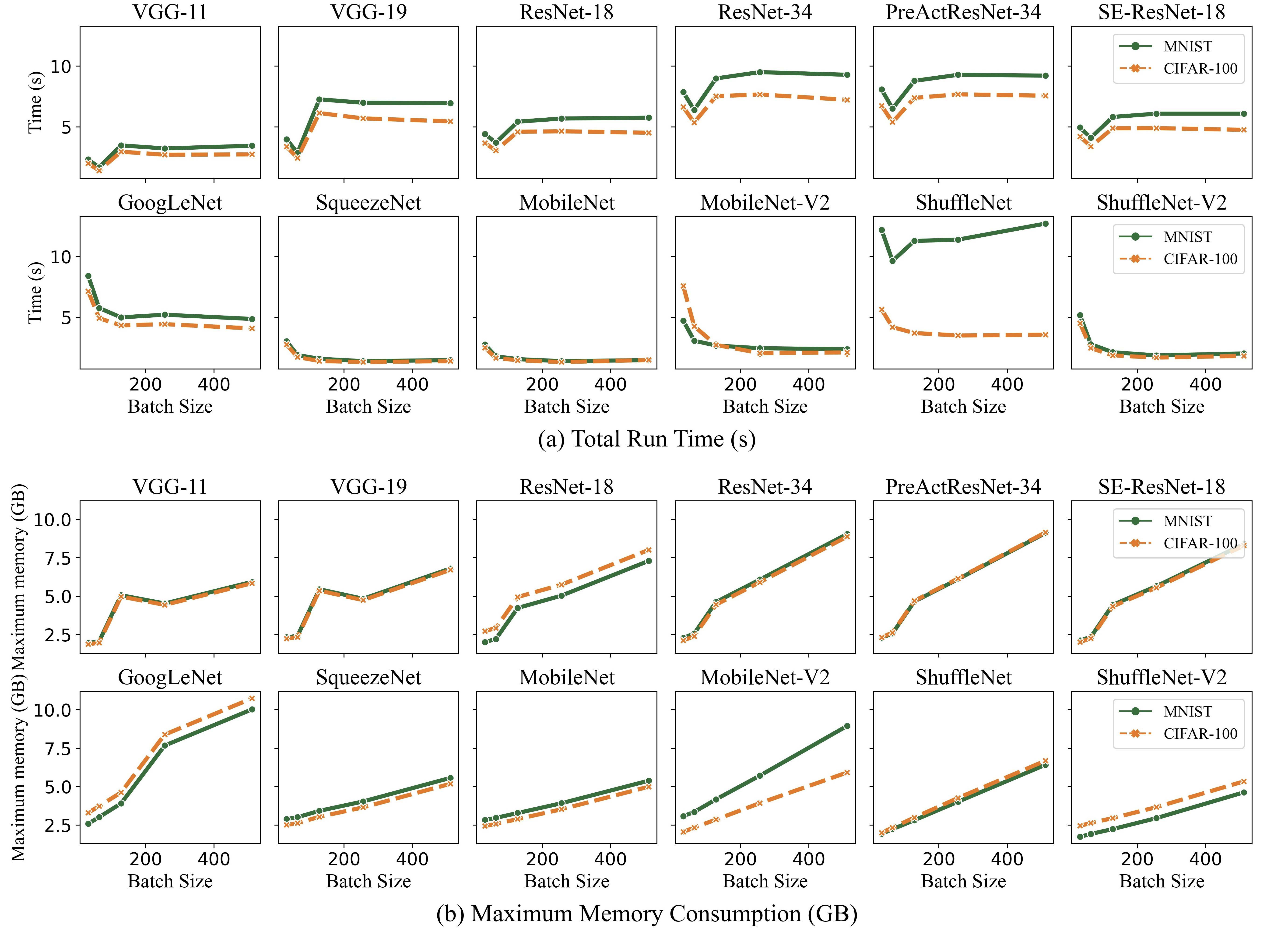}}
\caption{The correlation between batch size and the total run time (a) and maximum memory (b). Two datasets are collected with learning rate as 0.1, data size as 0.1, and epoch as 1.}
\label{allNet}
\end{center}
\end{figure}

The memory and time consumption of the same network is constant with different learning rates. In terms of epochs, the time and maximum memory space required are also relatively stable in general. Therefore, we fixed the learning rate to 0.1 and the epoch to 1 during data collection. By varying the data size and batch size on both MNIST and CIFAR-100, we tested the total run time and maximum memory consumption of typical deep neural networks. Since the total run time is linear to data size and the maximum GPU memory consumption is not 
sensitive to data size, we set data size to 0.1. Figure \ref{allNet} (a) shows the correlation between batch size and total run time, while Figure \ref{allNet} (b) shows the correlation between batch size and GPU memory consumption.

From Figure \ref{allNet}, we can observe that for the lightweight networks such as SqueezeNet, MobileNet, and ShuffleNet, which use $1\times$1 kernels, as the batch size increases, the GPU memory footprint increases and the overall run time reduces. It is reasonable because a larger batch size improves the utilization of highly parallel GPU devices, which also results in a higher memory footprint in convolutional computations.
Meanwhile, a significant fluctuation in maximum memory consumption accompanied by a corresponding fluctuation in the run time is observed for other networks.

\begin{figure}[ht]
\begin{center}
\centerline{\includegraphics[width=\columnwidth]{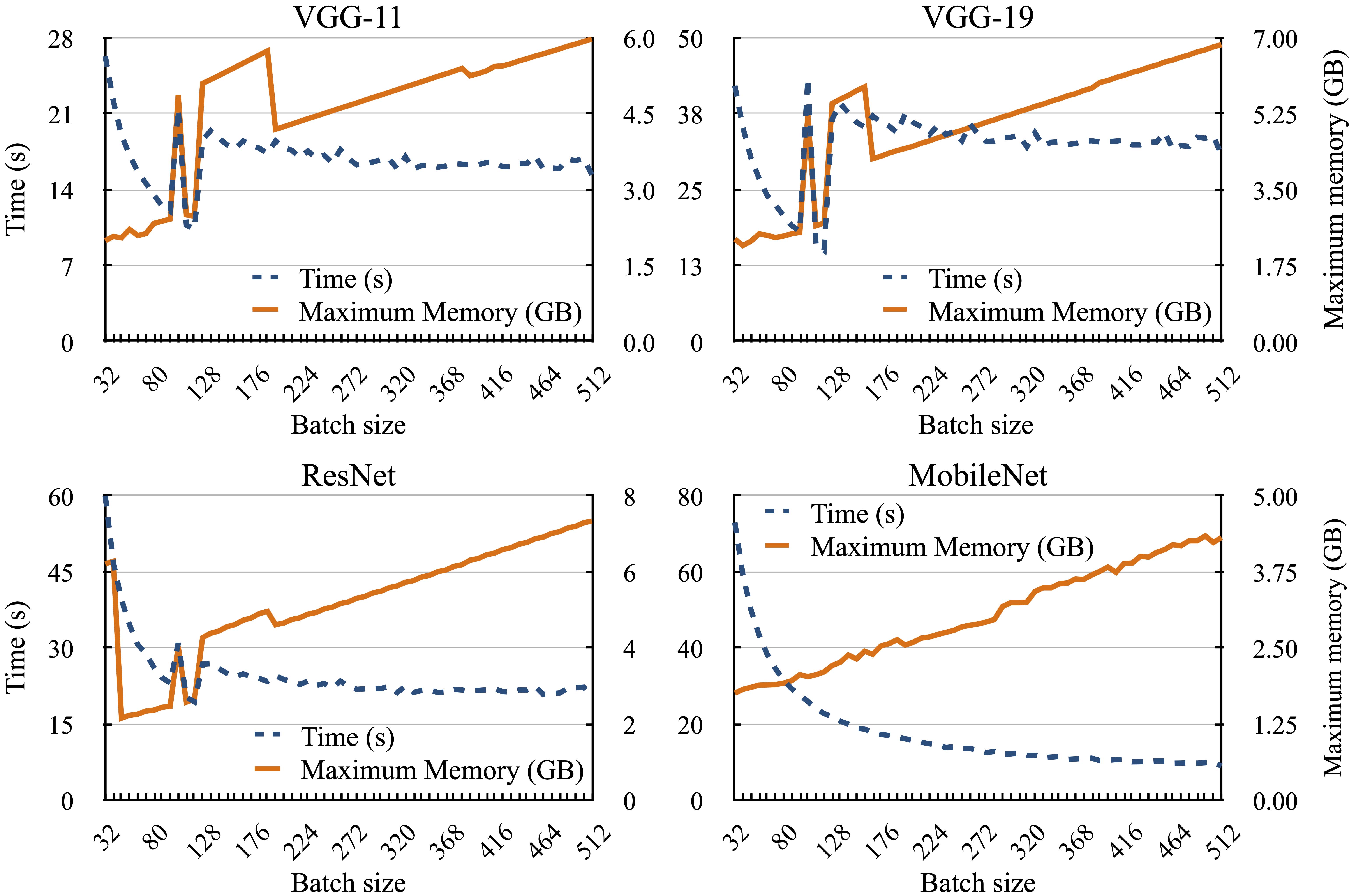}}
\caption{The total run time and the maximum memory consumption under different batch size with an interval of 2.}
\label{specNet}
\end{center}
\end{figure}

To confirm the occurrence of the fluctuation, we trained the deep neural network in different batch sizes with an interval of 2. As can be seen in Figure\ref{specNet}, the time consumption and the maximum memory consumption of networks without $1\times1$ convolution have undergone tremendous changes in some ranges between a batch size of 100 and 200.

Aiming to find out the reason for these fluctuations, we made a more detailed analysis for VGG-11, a network that fluctuates as the batch size increases, and MobileNet, a network that has stable performance, by extracting and analyzing the CuDNN logs generated in model training.

\begin{figure}[ht]
\begin{center}
\centerline{\includegraphics[width=\columnwidth]{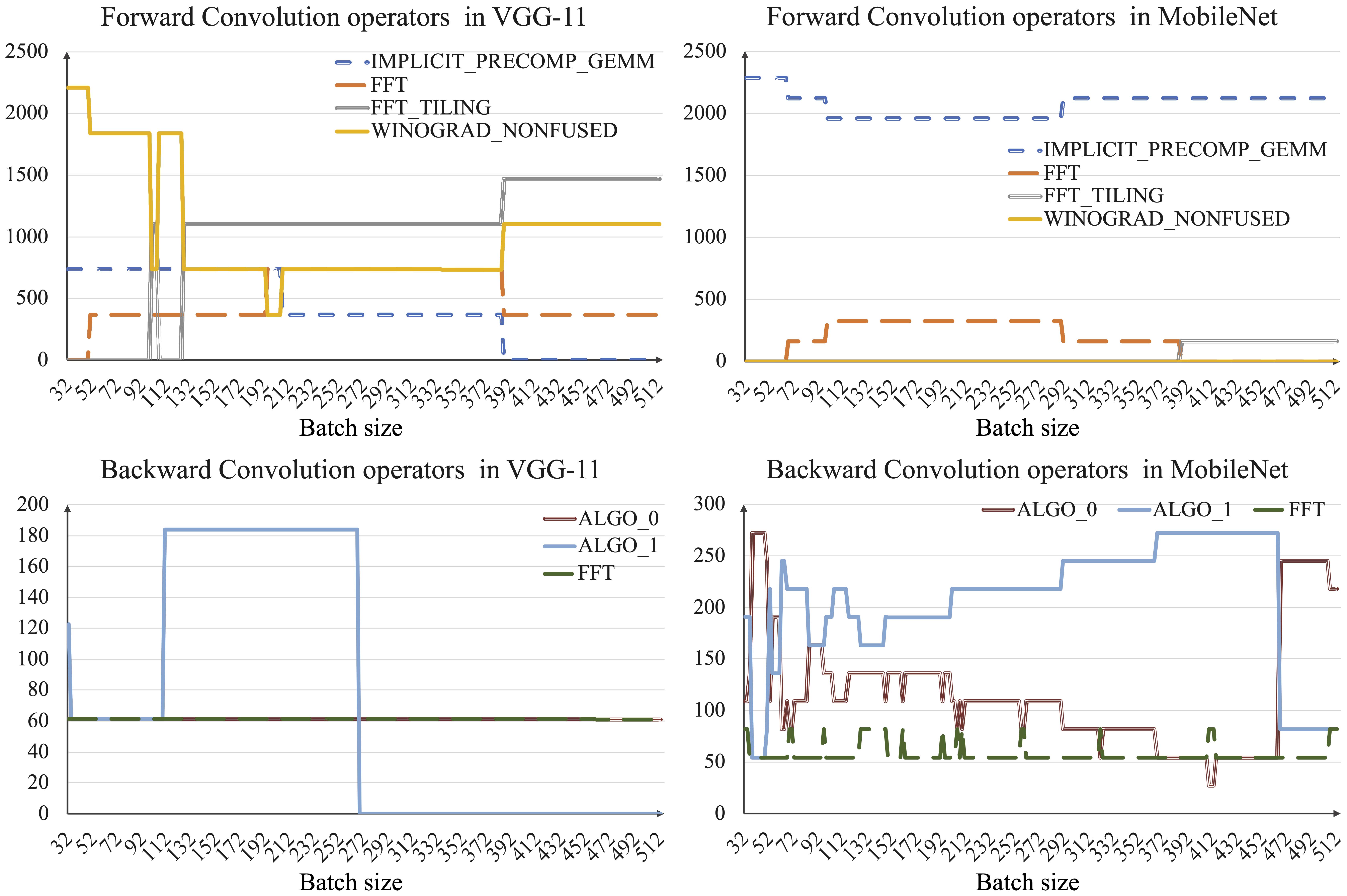}}
\caption{The number of convolution operators called by the networks as the batch size varies.}
\label{numAPI}
\end{center}
\end{figure}

Figure~\ref{numAPI} is the statistical results about the number of different convolution operators called by the network as the batch size varies. Since the variation of batch size causes different iteration numbers and tensor shapes, we normalize the total number of each convolutional kernel by dividing it over the sum of all kernels called. 
From Figure~\ref{numAPI}, we found that with the variations of the batch size, the number of convolutional operators called changes significantly.
Because of convolution algorithms have different time and space complexity, deep learning frameworks select convolution algorithms according to input tensor shape, used network structure, available memory at runtime.
What can be noticed is that MobileNet does not call WINOGRAD\_NONFUSED in forwarding passes, because it does not support $1\times1$ convolution. Therefore, networks that extensively use $1\time1$ convolution mostly calls GEMM, because it shows good performance with $1\time1$ filters.
For VGG-11, when the batch size is small, the network mostly calls WINOGRAD\_NONFUSED, which converts multiplication into addition operation and makes it suitable for small inputs. As the batch size increases, the number of FFT and FFT\_TILING called increases with the number of inputs, because the cost of transforming the inputs to the frequency domain and the result back to the time domain can be amortized by several convolutions.

\begin{figure}[ht]
\vskip 0.1in
\begin{center}
\centerline{\includegraphics[width=0.99\columnwidth]{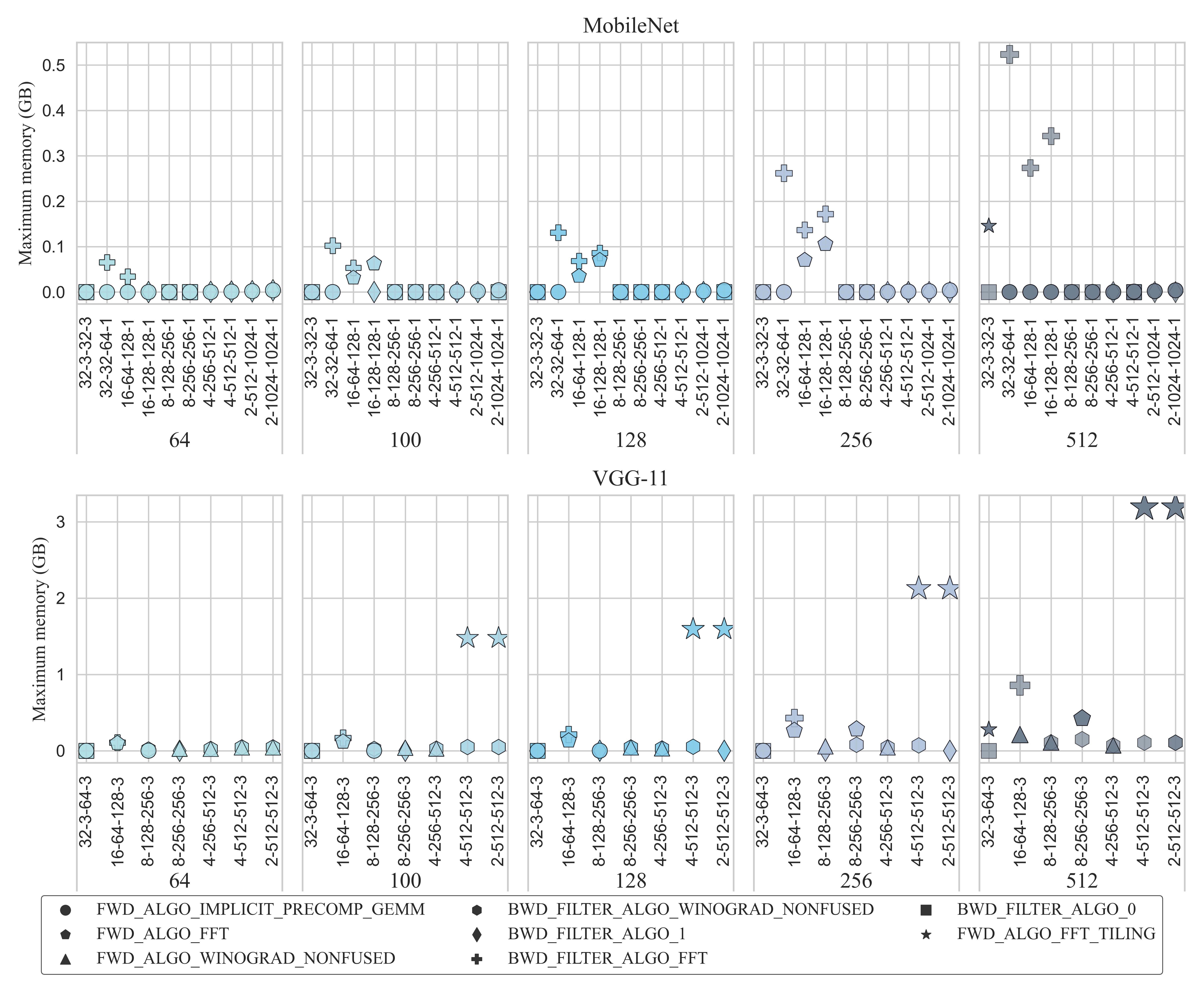}}
\caption{GPU memory consumption of convolution operators called under different convolution configurations. The X axis intervals indicate the batch size, the labels indicate [input X \&Y size]-[input depth]-[output depth]-[kernel X \&Y size].}
\label{conf}
\end{center}
\vskip -0.1in
\end{figure}

Since the number of convolution kernel calls does not explain the sudden change of memory footprint, we further collected the GPU memory consumption of each convolution operator called under different convolution configurations, as shown in Figure \ref{conf}. The convolution parameters include the input tensor, the convolution kernel and the output tensor. We found that for VGG-11, the peak memory consumption is achieved when FFT\_TILING is called, which is the reason for the fluctuation in memory consumption. The memory consumption of FFT\_TILING increases significantly when the number of input and output depth of the convolution kernel are large. In the case of MobileNet, the peak of the network is caused by FFT during backward passes.

In conclusion, the difference in convolution operator calls is the main factor for the variations of total run time and memory consumption, while the convolution operator calls are related to the network structures and the convolution configurations. For the lightweight network with a massive $1\times1$ convolutions, the total run time and the memory consumption are following the general regulation. For other networks, the run time and the memory consumption increase abruptly when the input and output depth of the convolution kernel is large because the FFT\_TILING is called.

However, convolution operator selection has uncertainties. VGG-11 also calls GEMM when the input tensor is small and stops using FFT when the batch size is large. MobileNet also calls FFT and FFT\_TILING in some configurations. Furthermore, the correlation of the configuration and the memory consumption of convolutional operators is not linear. FFT\_TILING requires a large memory space in some configurations. Due to the non-deterministic of convolutional operator selection, as well as other uncertainties, it is challenging to use an analytical model to directly calculate the total run time and memory consumption of deep neural models.

\section{Computational Cost Prediction}

\subsection{Overall Architecture}

Taking the memory prediction task as an example, the GPU memory occupied in the model training process is sensitive to parallel scheduling, memory management, convolutional algorithm, etc. It is challenging to build an analytical model to calculate the memory overhead. Unlike traditional analytical methods \cite{DNNMem2020}
, we employ a black-box approach and build a machine learning model for performance predicating. Besides, the main purpose of resources usage predicting is to guide the entire task scheduling before training. The prediction task itself has time and memory constraints and a lightweight network is preferred. We use the shallow machine learning models in an AutoML framework because they are fast and have small memory footprints.



\begin{figure}[ht]
\begin{center}
\centerline{\includegraphics[width=\columnwidth]{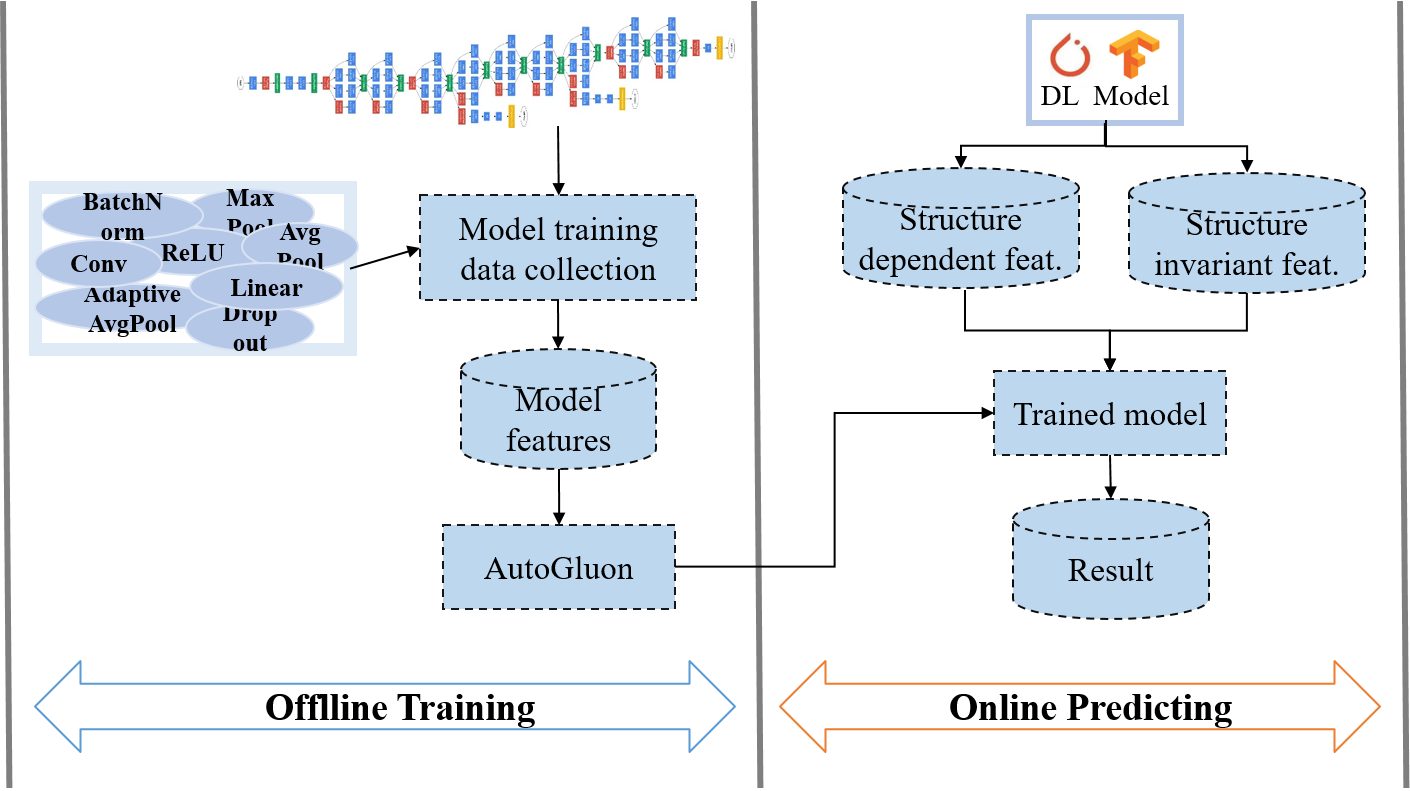}}
\caption{The overall architecture of our approach.}
\label{archi}
\end{center}
\end{figure}

Figure \ref{archi} illustrates the overall architecture of our approach. It has an offline training stage and an online predicting stage. The offline stage collects training data for popular deep learning models, such as GoogLeNet , ResNet, VGG and et al. In total, 29 networks were included and tens of thousands of configurations were tested. Not only that, to ensure the richness of the collected data, we also designed a random model generator and generated 5,500 test cases. The performance data were also collected with different configurations. Then, these features are fed into an AutoML framework for training. The model achieving the highest test accuracy is selected. Finally, we use the trained model to predict the performance of deep neural networks in the online prediction stage.

\subsection{Feature Engineering}
In this paper, features are classified into two categories. The first one is structure-independent features, and it intends to describe features that are not related to 
operators, such as the configuration of the model. The second category, called structure-dependent features, captures network structure characteristics, such as operators and their dependencies. 

\subsubsection{Structure-independent Features}

Structure-independent features describe the configuration of the model and are independent of the network structure, which are shown in Table \ref{features}. The first three features describe the shape of input tensors, and the last three features describe the overall magnitude of models. The remaining features are important parameters in model configuration. With a total of 9 features, we are able to predict the training performance of most models on GPU. This can also help users choose the appropriate model configurations.

\begin{table}[!ht]
    \caption{Structure-independent Features}
    \label{features}
    \centering
    \begin{tabular}{|l|p{0.3\textwidth}|}
    \hline
        Features &  Description  \\ \hline
        Batch Size & Number of samples for one iteration  \\ \hline
        Input Size & Height and width of input samples  \\ \hline
        Channel & Number of channels in input data  \\ 
        \hline
        Learning Rate & A tuning parameter in an optimization algorithm.  \\ 
        \hline
        Epoch & The number of epochs  \\ 
        \hline
        Optimizer  & A function or algorithm that modifies the attributes of the neural network  \\ \hline
        Layers & Number of layers in the model  \\ \hline
        FLOPs & Number of floating-point operations  \\ \hline
        Params & Number of parameters in the model  \\
        \hline
    \end{tabular}
\end{table}

\subsubsection{Structure-dependent Features}
 
Deep learning models can be formalized as tensor-oriented computation graphs, which is a directed acyclic graph (DAG).

\begin{equation}
  G = < {u_{i}}_{i=1}^{n} , e_{i,j} = ( u_{i} ,u_{j}) > 
\end{equation}

Each node $u_{i}$ represents a call to an operator in the model, such as Conv2D and BatchNorm2D. The inputs/outputs of the nodes are tensors (multidimensional numeric arrays). The directed edge $e_{i,j}$ indicates that the output tensor of the operator node $u_{i}$ is handed over to the operator node $u_{j}$ as the input tensor, and also specifies the execution dependency between the operators.


To extract the model structure in the computational graph, we design a novel and lightweight network architecture presentation technique, network Structural matrix (NSM), based on operators and their dependencies. In order to further compare and evaluate the network representation ability of NSM, we also include the representation method of graph embedding in this paper. 

\textbf{Network Structural Matrix (NSM)}. The purpose of NSM is to present the number of times each operator-pair appears in the computational graph. Each line or column is defined as an operator in networks, and each entry in the NSM is defined as the number of edges connecting the source operator in the row and the sink operator in the column. 

Let $S$ be a set of operators in graph $G$, $E$ be a topological ordering from the above graph edge ordering. DNNAbacus follows $E$ to traverse the computation graph $G$ sequentially. Assuming that the currently traversed edge is $e_{i,j}$, then the number in the cell$(i,j)$ is increased by 1. 





\begin{figure}[ht]
\vskip 0.0in
\begin{center}
\centerline{\includegraphics[width=0.8\columnwidth]{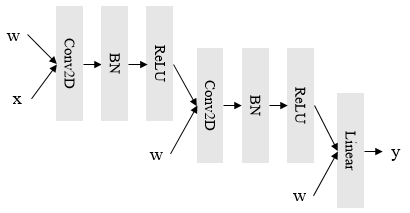}}
\caption{Computation graph of an example network}
\label{f_compute}
\end{center}
\vskip -0.2in
\end{figure}

\begin{figure}[ht]
\vskip 0.0in
\begin{center}
\centerline{\includegraphics[width=0.8\columnwidth]{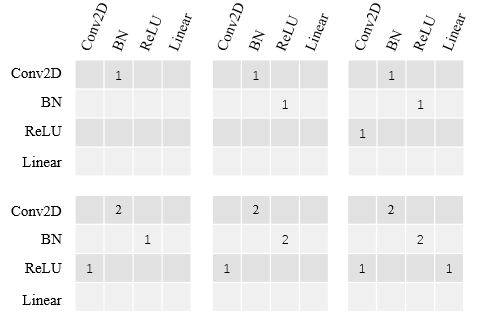}}
\caption{The construction process of the network structural matrix}
\label{example_arr}
\end{center}
\vskip -0.2in
\end{figure}

Taking a simple network as an example, the computation graph is shown in the Figure \ref{f_compute}, and the square nodes represent operators. This graph has seven operators of four types. The operators, from left to right, are numbered as $NO.i (i = 1, 2, 3, 4, 5, 6, 7)$. The type of each operator constitutes the rows and columns of the NSM. The construction of NSM for the given example is shown in Figure \ref{example_arr}.


Taking step 4 as an example, there is only one edge with node $NO.4$ as the starting point, and the output of this node is the input of the $BN$ operation. Therefore, the entry whose row and column are $Conv2D$ and $BN$ respectively is increased by 1. The NSM built from this  clearly shows the number of occurrences of operators, as well as the number of simultaneous occurrences of two operators. 

\textbf{Graph embedding} Graph embedding converts high-dimensional sparse graphs into low-dimensional, dense and continuous vectors, maintaining the structural properties of a graph to the greatest extent \cite{graphemded}.
 \cite{niepert2016learning} offer excellent performances in supervised learning tasks (e.g., graph classification), but acquiring large volume of labeled data is a challenge in itself. In order to compare with our lightweight presentation of NSM, we use an unsupervised model that succinctly captures general characteristics of the entire graph in the form of embeddings. We finally adopted \texttt{graph2vec} \cite{graph2vec} to generate graph embeddings for evaluation.
 

We use the \texttt{graph2vec} method to extract the rooted subgraphs $sg_{n}^{d}$ that possess nonlinear features in the computational graph $G$ and train the graph coding to obtain the embedding $E_{G}$ of $G$. After random initialization of $E_{G}$, we first extract the rooted subgraph $sg_{n}^{d}$ around each node in $G$ and use the SkipGram model to compute the target likelihood function $ J(E_{G})$. By introducing a learning rate $\alpha$, we use negative sampling to improve efficiency, which allows the $ J(E_{G})$ results to be updated in a faster iteration. After several iterations, we obtain the final $E_{G}$.
 
Similar to graph neural networks, graph embedding is also time-consuming in graph vectorization. However, NSM can be built in one-time scanning of the input graph.
 

\subsection{Model Training and Testing}

AutoML, which automatically builds machine learning models and tunes parameters, outperforms both in accuracy and feedback time. There are many popular AutoML frameworks available, such as AutoGluon \cite{autogluontabular}, AutoKeras \cite{autokeras}, and TPOT \cite{Olson2019}. AutoGluon is an open-source implementation of AutoML from Amazon that focuses on automated stack ensembling, deep learning, and real-world applications. Different from other AutoML frameworks, AutoGlion integrates multiple lightweight models and combines them into a new multi-layer hybrid model.

We collected 17,300 data points for the classic 29 models and 5,500 tests for randomly generated models on two servers in Table \ref{server}. Then, we shuffle all the data points in the first dataset and randomly select 70\% as the training set and the remaining 30\% as the test set. AutoGluon supports various types of shallow machine learning models including Random Forest, Gradient Boost Decision Tree, and Extra-Trees. We pick the model with the lowest mean relative error as the final performance model.


\section{Evaluation}


\subsection{Memory and Time Prediction}



\begin{figure}[ht]
\vskip 0.0in
\begin{center}
\centerline{\includegraphics[width=\columnwidth]{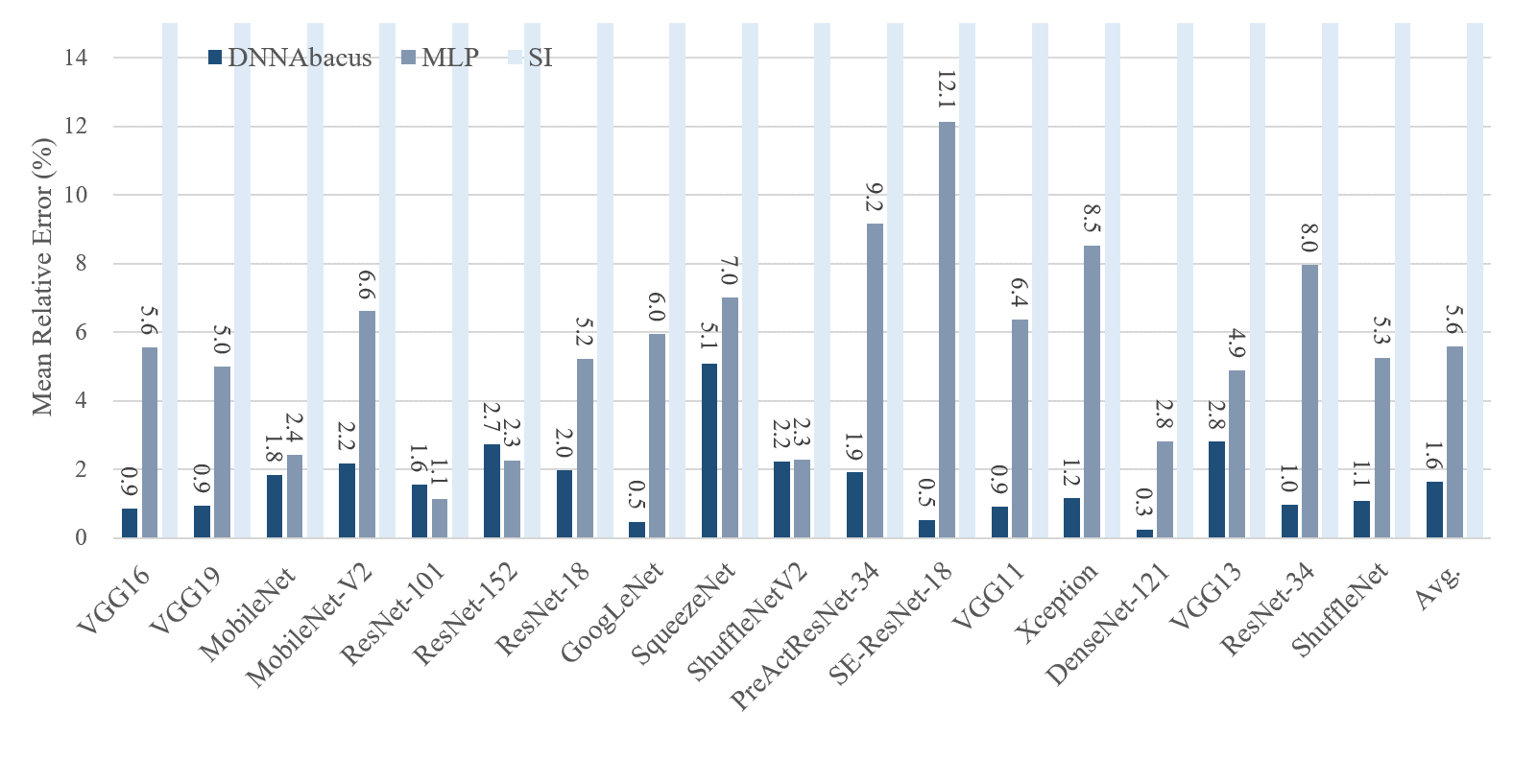}}
\caption{The MRE of memory prediction for PyTorch.}
\label{MRE-M-PT}
\end{center}
\vskip -0.2in
\end{figure}

\begin{figure}[ht]
\vskip 0.0in
\begin{center}
\centerline{\includegraphics[width=\columnwidth]{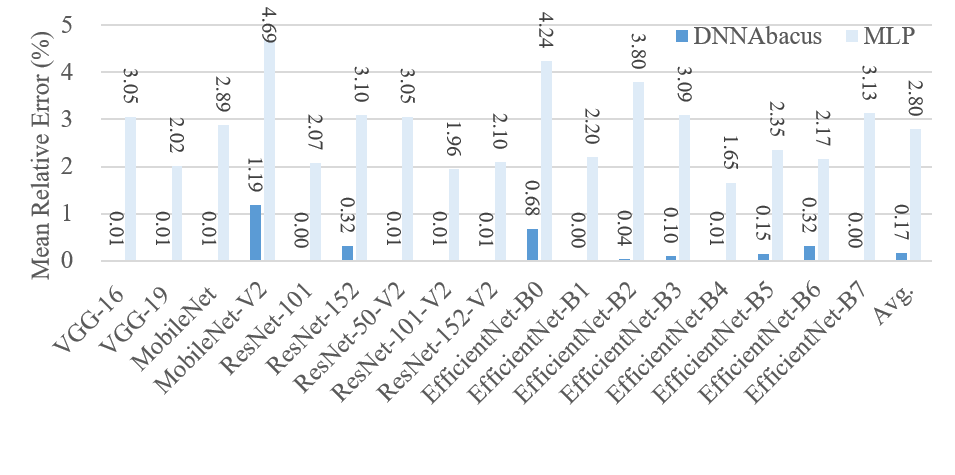}}
\caption{The MRE for memory prediction for TensorFlow.}
\label{MRE-M-TF}
\end{center}
\vskip -0.2in
\end{figure}

\begin{figure}[ht]
\vskip 0.0in
\begin{center}
\centerline{\includegraphics[width=\columnwidth]{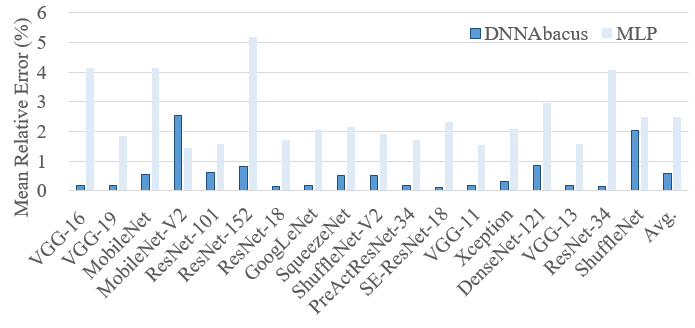}}
\caption{The MRE of time prediction for PyTorch.}
\label{MRE-T-PT}
\end{center}
\vskip -0.0in
\end{figure}

\begin{figure}[ht]
\vskip 0.0in
\begin{center}
\centerline{\includegraphics[width=\columnwidth]{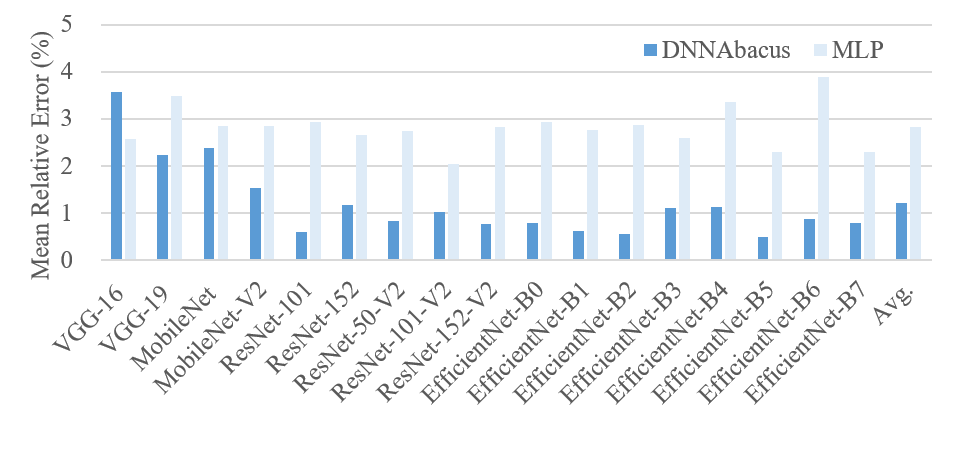}}
\caption{The MRE of time prediction for TensorFlow.}
\label{MRE-T-TF}
\end{center}
\vskip -0.4in
\end{figure}

In this section, we evaluate the overall performance of DNNAbacus in predicting GPU memory consumption and training time. Figure \ref{MRE-M-PT}-\ref{MRE-T-TF} show the overall results on 29 models, among them 17 models are implemented in TensorFlow, 18 models are implemented in PyTorch, and 6 models are used in both TensorFlow and PyTorch frameworks. 

The numerical result reveals that DNNAbacus has exceptional accurate predictions on a wide range of neural networks. For models implemented in PyTorch, the MRE of memory prediction ranges from 0.257\% to 5.096\%, with an average of 1.644\%; the MRE of time prediction falls between 0.121\% and 2.543\%, with an average of 0.573\%. For models implemented in TensorFlow, the MRE of memory prediction ranges from 0.001\% to 1.185\%, with an average of 0.167\%; the MRE of time prediction falls between 0.502\% and 3.581\%, with an average of 1.209\%. 

Two comparative methods are used to evaluate the performance of DNNAbacus. One is shape inference~\cite{si2020}, aiming to infer the consumption with the shape of operator inputs, outputs, and weights. The MRE of shape inference reaches 46.8\% in memory prediction for PyTorch, which is too high to be captured in Figure \ref{MRE-M-PT}. Similar results take place in other three comparison experiments. The other comparison method is using MLP to model the resource consumption. The multi-layer network is derived from recent works ~\cite{perfnet2020} and ~\cite{wu2020methods}. For models implemented in PyTorch, the MRE of memory prediction is 5.6\% on average, which is much higher than DNNAbacus. Figure \ref{MRE-M-PT}-\ref{MRE-T-TF} show the comparison in detail, reflecting the superiority of DNNAbacus.




\begin{figure}[ht]
\vskip 0.2in
\begin{center}
\centerline{\includegraphics[width=\columnwidth]{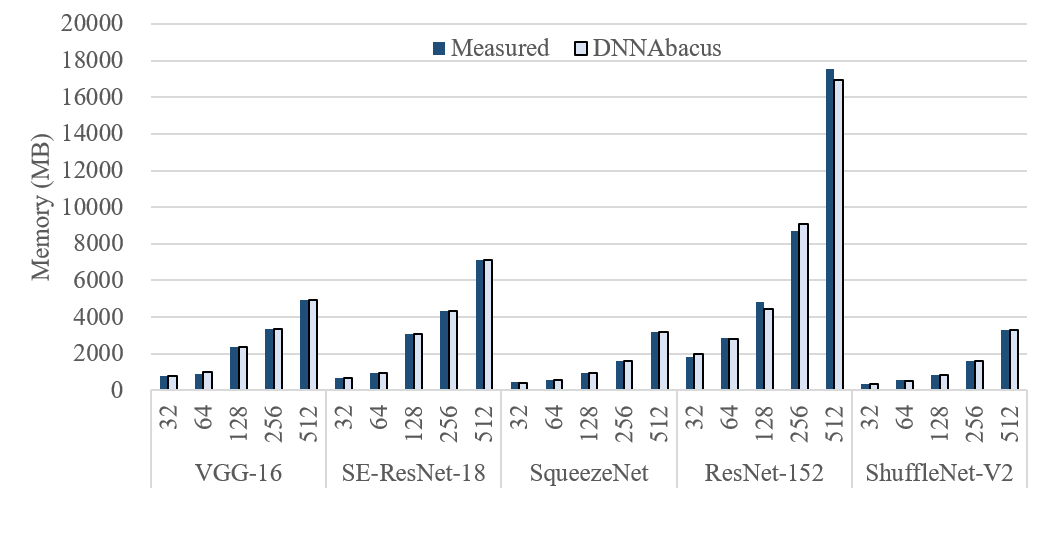}}
\caption{Maximum GPU memory predicting of five models with different batch size.}
\label{e2m}
\end{center}
\vskip -0.2in
\end{figure}

To further evaluate DNNAbacus, we experiment with the five deep learning models implemented on PyTorch, using five different batch size ranging from 32 to 512. As shown in Figure \ref{e2m}, the MRE of memory prediction across all VGG-16, SE-ResNet-18, SqueezeNet, ResNet-152, and ShuffleNet-V2 configurations are 3.46\%, 0.27\%, 1.46\%, 5.68\%, and 1.80\% on average, respectively. The results demonstrate the effectiveness of DNNAbacus under a wide range of different configurations.


\subsection{Performance of Unseen Networks}
\begin{figure}[ht]
\vskip 0.0in
\begin{center}
\centerline{\includegraphics[width=\columnwidth]{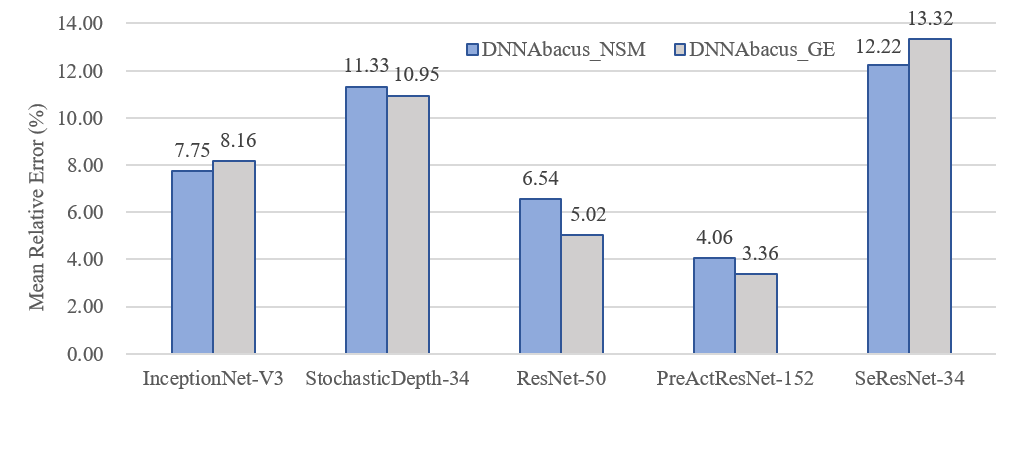}}
\caption{The evaluation results of unseen models. DNNAbacus\_NSM and DNNAbacus\_GE represent structure matrix and graph embedding based network vectorization, respectively.}
\label{e3m}
\end{center}
\vskip -0.4in
\end{figure}

In this section, we evaluate the generality of DNNAbacus, which is crucial for the practicability of our tool. A challenge for real-time prediction system is estimating the resource consumption of unseen deep learning models. We construct the testing dataset of different model configurations for InceptionNet-V3, StochasticDepth-34, ResNet-50, PreActResNet-152 and SeResNet-34. These five deep learning models are not included in the training dataset. Figure \ref{e3m} shows the maximum MRE of five unseen models using structure matrix based DNNAbacus and graph embedding based DNNAbacus. The results show that MRE of structured matrix based DNNAbacus and graph embedding based DNNAbacus are 8.38\% and 8.16\%, revealing the reliable generality of DNNAbacus. DNNAbacus not only performs well on unseen models, but also has good generality on unseen model inputs, with an average relative error of 3.65\%.

\subsection{Scheduling Application}


In this section, we use the predicted time and memory consumption to schedule deep learning training jobs on servers. The principle is cutting down the total training time without OOM failures. We use three scheduling plans to arrange 20 deep learning training jobs with different neural networks on the two servers described in Section 2.2. The first scheduling plan is the optimal scheme, with a total training time of 783.2s. The second scheduling plan is simply arranging the training jobs at random. The average total training time is 990.1s in 100 repeated trials. The third scheduling plan is using genetic algorithm to optimize the scheduling. We use gene coding, i.e. a 0-1 string with a length of 20 to represent the scheduling scheme. We denote 0 as training the network on Machine 1, 1 as training the network on Machine 2, and set the initial population number as 20. The fitness function is the training time of an individual,  and 20 individuals with the best fitness function are selected as the parents of next population. Genetic algorithm with our predicated memory and time reaches a total training time of 783.2s after 20 generations, which is the same as the optimal scheme and 20.9\% shorter than random planning on average.

\begin{figure}[ht]
\begin{center}
\centerline{\includegraphics[width=0.8\columnwidth]{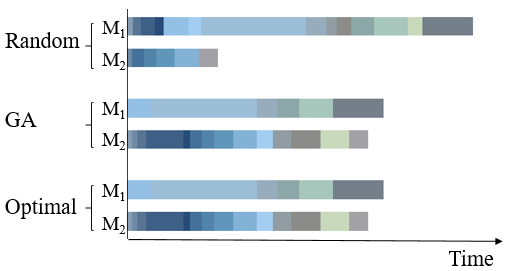}}
\caption{Task scheduling diagram}
\label{diaodu}
\end{center}
\vskip -0.2in
\end{figure}

We present the scheduling scheme of three methods in Figure \ref{diaodu}. The plot shows these 20 training jobs in different colors where the length is proportional to the execution time. It is clearly noticed that the genetic algorithm reaches the optimal planning, thus verifying the reliable performance. 


\section{Related Work}


\textbf{Time prediction} Many researchers focus on predicting the execution time of deep learning jobs~\cite{Qi2017PaleoAP, prediction18, adlers2018prediction,  layer2019, yang2021perfestimator, ponomarev2021latency}. Paleo~\cite{Qi2017PaleoAP} builds an analytical model to estimate the time consumption of deep learning jobs. By modeling computation time and communication time respectively, Paleo estimates the total elapsed time under different network structures, communication strategies and hardware configurations. Justus et al.~\cite{prediction18} builds a model to predict the execution time for convolution layers, thus putting together to estimate the total completion time of the entire network. One limitation is that only VGG-16 is investigated in this work. Gianniti et al.~\cite{layer2019} also aims at predicting the execution time of each layer. The approach is fitting a linear model between the execution time and the complexity of each layer, where the data is collected manually on GPU. Three neural networks (LeNet, GoogLeNet and VGG-16) are evaluated in their experiments. 

These analytical approaches rely on the selected features, hence great hand-draft efforts are required. Some recent works~\cite{perfnet2020,Habitat21,PerfNet22021} use neural networks to predict time cost automatically. Wang et al.~\cite{perfnet2020} proposes a four-layer regression model, called PerfNet, to predict the inference time of neural networks. As reported in the numerical results concerning three networks, PerfNet largely outperforms the comparative methods, including ~\cite{prediction18}. Their following up work ~\cite{PerfNet22021} proposes a platform-aware performance model called PerfNet-V2, which can be applied to predict both inference and training time. PerfNet-V2 adds one-dimensional convolution kernels into the prediction model, thus reflecting a further decline on the relative error. PerfNet-V2 is also reliable on predicting the time consumption of unseen deep learning models, indicating another advantage over previous methods.

\textbf{Memory Prediction} Shape inference~\cite{si2020} is an applicable tool for memory prediction. Using shape inference, the size of weights, input and output tensors can be discovered from the computation graph. But these parameters only make up part of the memory consumption, leading to the underestimation of memory cost. Wu et al.~\cite{wu2020methods} uses classical machine learning methods to predict the memory demand. As reported in the results, the prediction is not accurate enough.

Some recent works manage to quantify memory consumption more systematically. Gao et al.~\cite{DNNMem2020} proposes DNNMem as an estimation tool for GPU memory consumption of deep learning models. The prediction is carried out based on computation graphs and the configurations of deep learning jobs. Through evaluation on three popular deep learning frameworks (TensorFlow, PyTorch, and MXNet) with five neural networks (VGG-16, ResNet-50, Inception-V3, LSTM and BERT), the numerical results show a competitive performance on DNNMem. Their following up work~\cite{gao2021runtime} proposes DNNPerf, a novel algorithm to predict the memory consumption and time cost of deep learning models using graph neural networks. Their experimental results on ten neural networks reveal that DNNPerf significantly outperforms the competitive methods. Same as PerfNet-V2, DNNPerf performs well on unseen deep learning models. 

\textbf{Workload Scheduling} There are some works considering deep learning workloads scheduling on GPU clusters~\cite{peng2018optimus, xiao2018gandiva, gu2019tiresias, le2020allox, hu2021characterization}.  Gandiva~\cite{xiao2018gandiva} utilizes domain-specific knowledge to lower the latency. Tiresias ~\cite{gu2019tiresias} uses the Gittins index as the job priority. Optimus~\cite{peng2018optimus} trains a simple but effective model online to minimize job completion time. AlloX ~\cite{le2020allox} uses min-cost bipartite matching to solve scheduling problems and provides a long-term dynamic allocation.  ~\cite{hu2021characterization} collects deep learning job traces from a datacenter called Helios to extract the workload patterns, thus making the resource demand predictable. Then an algorithm derived from Shortest-Service-First is carried out to optimize the job completion time.

\section{Conclusion}
This paper investigates the computational resource predicating problem of deep neural networks. We firstly profiled 29 representative deep neural models on two popular deep learning frameworks, TensorFlow and PyTorch. Then we deeply analyze the impact of hyperparameters on model training time and maximum GPU memory demands. Based on our analysis, accurate predicting models are presented for training time and memory predicating. Different from existing work, we take model training on frameworks as a black box to tolerate complicated memory management and parallel scheduling in frameworks, operator combination and interactions in networks. Evaluation results show that our predicating models are accurate compared with shape inference and MLP based models. Our approach is also adaptive and useful to unseen networks. We apply the models in a task scheduling problem of 20 networks based on a genetic algorithm. Results show that it achieves optimal results after 20 iterations.


\bibliographystyle{ACM-Reference-Format}
\bibliography{example_paper}

\end{document}